%% file: main.tex
\documentclass[10pt,twocolumn,letterpaper]{article}

\usepackage[pagenumbers]{cvpr} % To force page numbers, e.g. for an arXiv version
\usepackage{threeparttable}
\usepackage{array}
\usepackage{booktabs}
\usepackage[table]{xcolor}
\usepackage{graphicx}
\usepackage{multirow}
\usepackage{tikz}
\usepackage[table]{xcolor}
\usepackage{graphicx}
\usepackage{longtable}
\usepackage[misc]{ifsym} %for \Letter
\definecolor{myrowgray}{RGB}{240,240,240}

\usepackage{fix-cm}
\usepackage[accsupp]{axessibility}

\usepackage{pifont}
\newcommand*\circled[1]{\tikz[baseline=(char.base)]{
            \node[shape=circle,fill=black,text=white,draw,inner sep=0.5pt] (char) {#1};}}

\input{preamble}
\definecolor{cvprblue}{rgb}{0.21,0.49,0.74}
\usepackage[pagebackref,breaklinks,colorlinks,allcolors=cvprblue]{hyperref}

\definecolor{darkgray}{gray}{0.3}
\newcommand{\psddataset}{{CreativePSD}\xspace}
\newcommand{\plannername}{{GraphicPlanner}\xspace}
\newcommand{\systemname}{{PSDesigner}\xspace}
\newcommand{\collectorname}{{AssetCollector}\xspace}
\newcommand{\modeonename}{$\mathcal{X}_\text{gen}$\xspace}
\newcommand{\modetwoname}{$\mathcal{X}_\text{edt}$\xspace}
\newcommand{\executorname}{{ToolExecutor}\xspace}
\newcommand{\myparagraph}[1]{{\vspace{.3em} \noindent \bf #1}}
\def\paperID{2761} % *** Enter the Paper ID here
\def\confName{CVPR}
\def\confYear{2026}

\title{\systemname: Automated Graphic Design with a Human-Like Creative Workflow}

\author{
Xincheng Shuai$^1$\footnotemark[1]
\quad
Song Tang$^1$\footnotemark[1]
\quad
Yutong Huang$^1$
\quad
Henghui Ding$^1$~\!$^{\textrm{\Letter}}$
\quad
Dacheng Tao$^2$
\vspace{.6mm}
\\
{\fontsize{11}{11}\selectfont $^1$Institute of Big Data, College of Computer Science and Artificial Intelligence, Fudan University, China}\\
{\fontsize{11}{11}\selectfont $^2$Generative AI Lab, College of Computing and Data Science, Nanyang Technological University, Singapore}
\\
{\tt\footnotesize henghui.ding@gmail.com\quad dacheng.tao@gmail.com}
\\
\href{https://henghuiding.com/PSDesigner/}{https://henghuiding.com/PSDesigner/}
}

\begin{document}
\twocolumn[{%
\renewcommand\twocolumn[1][]{#1}%
% \maketitle
\maketitle
% \vspace{-7.6mm}
\begin{center}
\centering
\captionsetup{type=figure}
\includegraphics[width=0.98\linewidth]{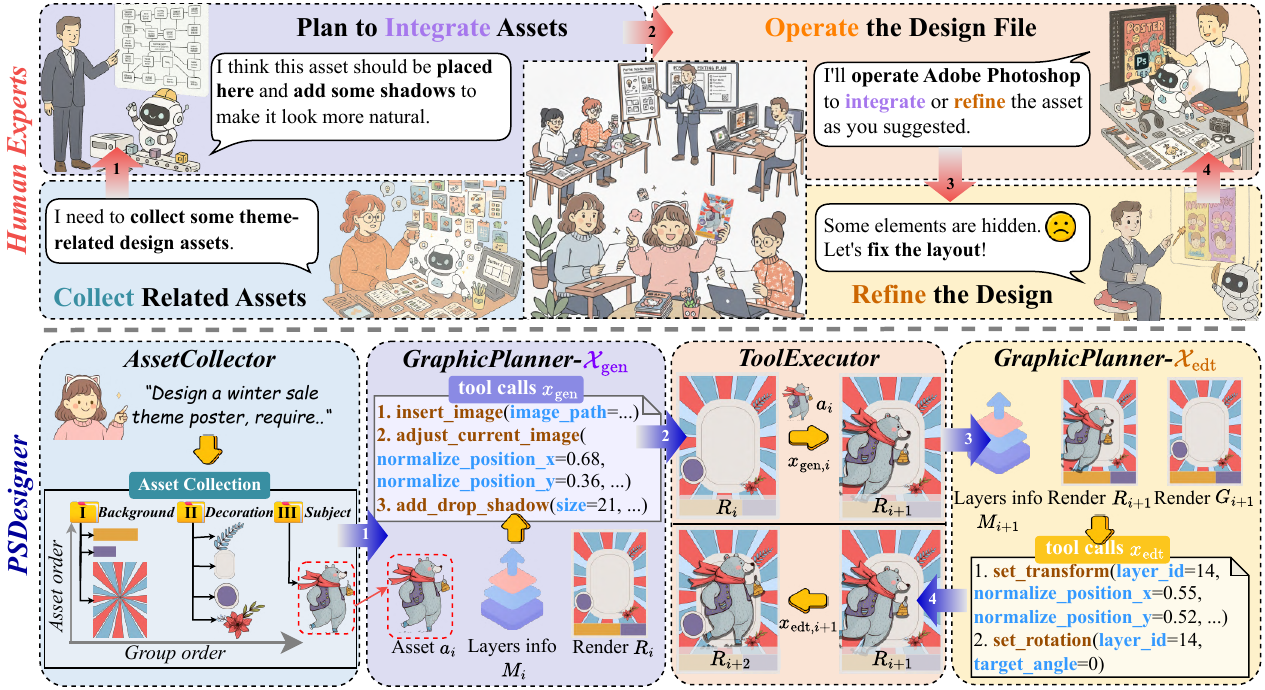}
% \vspace{-3.16mm}
\vspace{-2.16mm}
\captionof{figure}{The figure illustrates the high similarity between the graphic design workflows of human experts (top) and \systemname (bottom). They begin by collecting theme-related assets based on the user instructions. Next, they iteratively integrate these assets, where a bottom-up traversal is performed on the nested hierarchy, first at the group level and then at the asset level. In particular, each step consists of planning (\scalebox{0.75}{\protect \circled{1}}) and inserting (\scalebox{0.75}{\protect \circled{2}}) the current asset, then identifying deficiencies (\scalebox{0.75}{\protect \circled{3}}) and performing refinements (\scalebox{0.75}{\protect \circled{4}}). The above steps are repeated until all assets are integrated into the design file.}
\vspace{2.96mm}
\label{fig:teaser}
\end{center}
}]
\maketitle

\renewcommand{\thefootnote}{\fnsymbol{footnote}}
\footnotetext[1]{Equal contribution.}
\footnotetext[0]{${\textrm{\Letter}}$ Corresponding author (henghui.ding@gmail.com).}

\input{sec/0_abstract}
\input{sec/1_introduction}

\input{sec/2_related_works}

\input{sec/3_dataset}
\input{sec/3_method}
\input{sec/4_experiments}
\input{sec/5_conclusion}

{
    \small
    \bibliographystyle{unsrt}
    \bibliography{main}
}

\end{document}

%% file: sec/0_abstract.tex
\begin{abstract}
Graphic design is a creative and innovative process that plays a crucial role in applications such as e-commerce and advertising. However, developing an automated design system that can faithfully translate user intentions into editable design files remains an open challenge. Although recent studies have leveraged powerful text-to-image models and MLLMs to assist graphic design, they typically simplify professional workflows, resulting in limited flexibility and intuitiveness. To address these limitations, we propose \textbf{\systemname}, an automated graphic design system that emulates the creative workflow of human designers. Building upon multiple specialized components, {\systemname} collects theme-related assets based on user instructions, and autonomously infers and executes tool calls to manipulate design files, such as integrating new assets or refining inferior elements. To endow the system with strong tool-use capabilities, we construct a design dataset, \textbf{\psddataset}, which contains a large amount of high-quality PSD design files annotated with operation traces across a wide range of design scenarios and artistic styles, enabling models to learn expert design procedures. 
Extensive experiments demonstrate that {\systemname} outperforms existing methods across diverse graphic design tasks, empowering non-specialists to conveniently create production-quality designs.

\end{abstract}

%% file: sec/1_introduction.tex
\vspace{-6mm}
\section{Introduction}
\label{sec:intro}
\vspace{1mm}
Graphic design conveys rich visual and textual information, playing a significant role in fields like advertising, branding, and marketing, \etc. Traditional workflows require professional human designers to manually manipulate visual elements using sophisticated tools, such as Adobe Photoshop. However, this process demands substantial expertise and human effort, posing a significant challenge for non-specialists. 
% Since graphic design is a highly creative and innovative process,
Therefore, developing an automated design system remains an important and unsolved challenge.

\begin{figure}[t]
    \centering
    \includegraphics[width=1\linewidth]{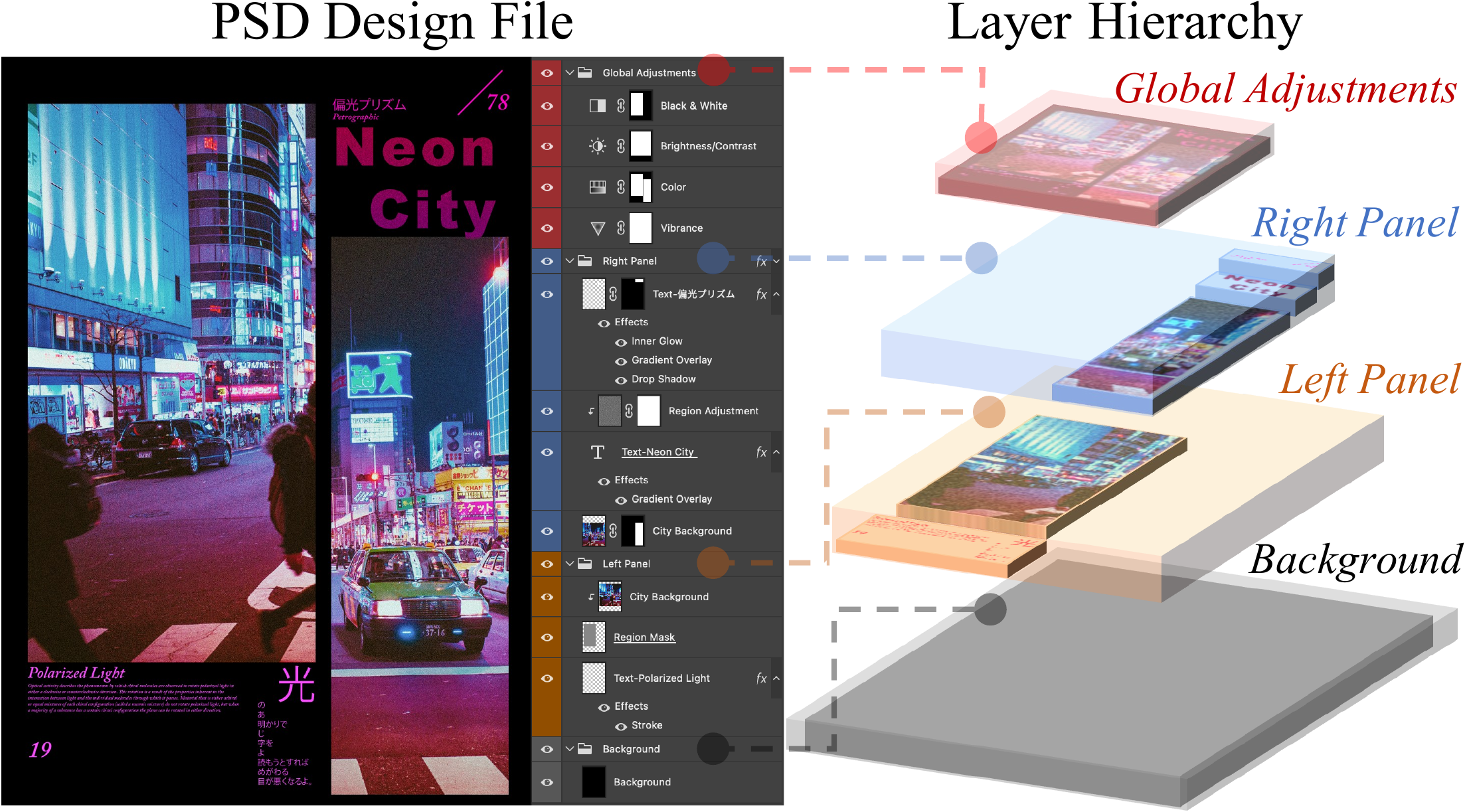}
    \vspace{-7mm}
    \caption{The typical layer hierarchy in PSD (Adobe Photoshop Document) files, where the layers used to compose the same visual concept (\eg, ``Left Panel'') are grouped together.}
    \vspace{-4mm}
   \label{fig:psd}
\end{figure}

We begin by examining the typical graphic design workflows of human experts. As shown at the top of \cref{fig:teaser}, they first collect theme-relevant assets, which are then incorporated into the design file step by step. In addition, designers also refine the inferior elements in each step until satisfaction.
\cref{fig:psd} shows an example of a PSD (Adobe Photoshop Document) file, where layers associated with the same visual concept are grouped together. In particular, designers achieve visually appealing design by configuring complex attributes of each layer, such as effects and masks.

Recently, a growing group of studies has leveraged machine learning~\cite{jyothi2019layoutvae,yu2024layoutdetr,hsu2025postero,tang2023layoutnuwa,lin2025elements} to assist graphic design. However, they have greatly simplified the process compared to the creative human workflow described above. One line of research~\cite{chen2024textdiffuser,chen2023textdiffuser,tuo2023anytext,yang2023glyphcontrol} employs text-to-image (T2I) models~\cite{gao2025seedream,podell2023sdxl,esser2024scaling,tuo2024anytext2,black2024flux,shuai2025free,li2025anyi2v,qin2025scenedesigner,shuai2025free2,shuai2024survey} to create high-quality design images using well-designed user prompts. However, they struggle to create accurate texts, resulting in missing or extraneous characters. Moreover, the generated images are non-editable, hindering them from adding customized elements or refining the content. 

To overcome these issues, other methods~\cite{gao2023textpainter,xu2023unsupervised,seol2024posterllama,shi2025layoutcot,hsu2025scan,li2019layoutgan,kikuchi2021constrained,jia2023cole,zhang2025creatiposter,chen2025posta,kikuchi2025multimodal,li2023planning} leverage Multimodal Large Language Models (MLLMs)~\cite{wang2024qwen2,chen2024internvl} and directly create the editable design files (\eg, JSON), encompassing the attributes of each layer, such as position and size. Most methods group layers into predefined logical categories, \eg, underlay and text, and jointly predict all layer attributes within each group. For example, LaDeCo~\cite{lin2025elements} first infers the attributes of all image layers and subsequently those of text layers. Nevertheless, they face the following challenges. \textbf{1). Non-intuitive design process.} Compared to the category-based grouping strategy, collaboratively inferring the attributes for layers grouped by visual concepts is more intuitive since they are more visually related, as indicated in \cref{fig:psd}. In addition, human designers typically configure and refine the layers progressively based on the current visual outcome, whereas predicting all layer attributes at once further reduces both flexibility and intuitiveness.  \textbf{2) Limited design operations.} Existing methods have only explored simple design scenarios, constrained by shallow layer hierarchies and limited layer\&attribute types. However, the product-level designs are far more complex, as shown in \cref{fig:psd}.
% the design incorporates adjustment layers to modify the visual style.

\begin{figure*}[!t]
    \centering
    \includegraphics[width=1\linewidth]{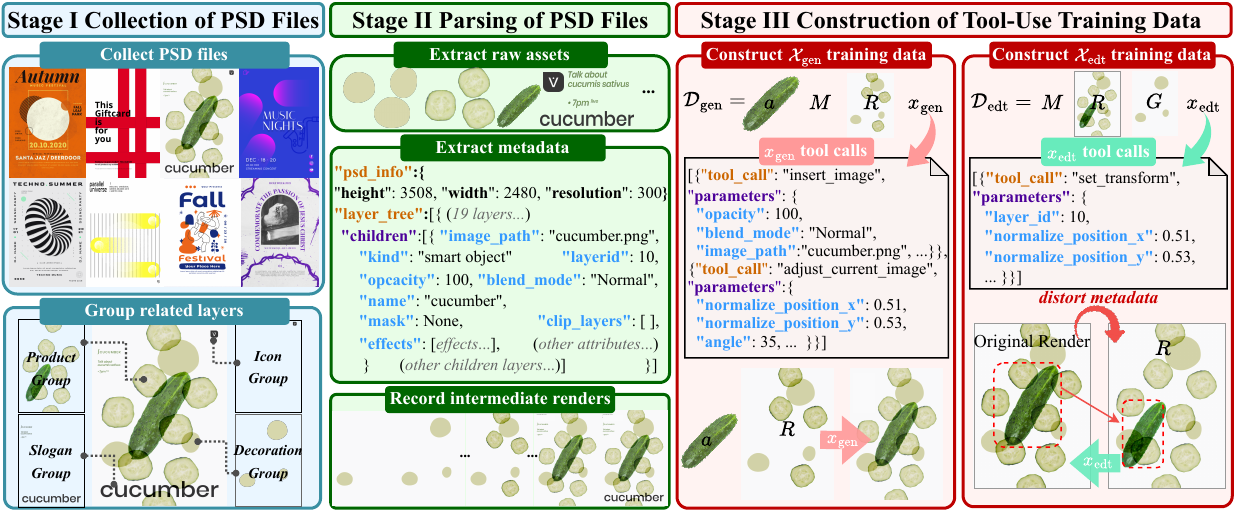}
    \vspace{-6.6mm}
    \caption{The three-stage construction pipeline of the proposed design dataset \psddataset. We first collect high-quality PSD files from the internet and paid data, while grouping the layers based on their underlying visual concepts. Then, we parse the PSD files and extract essential information, such as raw assets, metadata, and intermediate renders. Finally, we use the extracted data to construct the training data for \modeonename and \modetwoname modes of \plannername.}
   \label{fig:dataset_pipeline}
   \vspace{-2mm}
\end{figure*}

To address these challenges, we propose \textbf{\textit{\systemname}}, an automated graphic design system with a human-like creative workflow, enabling users to create visually appealing designs. As shown at the bottom of \cref{fig:teaser}, it integrates multiple modules to emulate the human experts. First, \collectorname collects the related assets for each identified visual concept from the user instruction, which are then iteratively incorporated into the design file. In each iteration, \plannername first predicts tool calls based on the current design to harmoniously integrate the asset. Then, it further infers tool calls to refine the inferior layers within the same group, thereby composing a coherent visual concept. These tool calls are performed by \executorname to directly operate the design file. Building upon this framework, our system is beneficial for addressing the first challenge mentioned above. To endow \plannername with strong tool-use capabilities and to tackle the second challenge, we construct \textbf{\textit{\psddataset}}, which contains a large number of high-quality PSD files annotated with operation traces, covering a wide range of design scenarios and artistic styles. To the best of our knowledge, \psddataset is the first design dataset based on the PSD format, facilitating the model to learn expert design procedures. Our contributions are:
\begin{enumerate}
    \item We propose \textit{\textbf{\systemname}}, an automated graphic design system with a creative design process, significantly simplifying the workflows for non-specialists.
    \item We introduce \textit{\textbf{\psddataset}}, the first design dataset based on the PSD format with operation traces, endowing the model with a powerful tool-use capability.
    \item  Extensive experiments demonstrate that \systemname outperforms other methods across diverse design tasks.  
\end{enumerate}

%% file: sec/2_related_works.tex
\section{Related Works}
\label{sec:related_works}

\myparagraph{Visual Text Rendering.}
Recently, a group of studies~\cite{wang2025uniglyph,wang2025designdiffusion,li2024joytype,tuo2024anytext2,ma2025glyphdraw2,ma2023glyphdraw,zhang2025creatidesign,gao2025postermaker} has made efforts to enhance the text rendering capabilities of T2I models~\cite{rombach2022high,black2024flux,podell2023sdxl}. Some of these methods, like Glyph-Byt5~\cite{liu2024glyph} and Seedream~\cite{gao2025seedream}, leverage character-level~\cite{liu2024glyphv2,liu2024glyph} or multilingual text encoders~\cite{gao2025seedream} to encode the text to be rendered. Other approaches, like AnyText~\cite{tuo2023anytext} and GlyphDraw~\cite{ma2023glyphdraw}, take the rendered glyph images as conditions to better generate out-of-vocabulary characters. However, these methods struggle to generate accurate texts, resulting in missing or extraneous characters. Furthermore, they primarily produce non-editable raster images. Consequently, users are unable to refine the results or incorporate their own customized assets, limiting the applications of these methods in practical design workflows.

\myparagraph{Automated Graphic Design System.}
Recently, a growing group of studies has integrated MLLMs~\cite{wang2024qwen2,chen2024internvl} to assist the design process. However, most of these approaches have greatly simplified this procedure compared with human experts. The early works~\cite{cao2022geometry,horita2024retrieval,zhang2023layoutdiffusion,gupta2021layouttransformer,zhou2022composition,kong2022blt,yamaguchi2021canvasvae,forouzandehmehr2025cal,wu2025layoutrag,inoue2023towards,cheng2025graphic,yang2024posterllava,guerreiro2024layoutflow,li2023relation,shabani2024visual,lin2023layoutprompter,chen2024towards,inoue2023layoutdm,hui2023unifying,hsu2023posterlayout,gao2023textpainter,xu2023unsupervised,seol2024posterllama,shi2025layoutcot,hsu2025scan,li2019layoutgan,kikuchi2021constrained,jyothi2019layoutvae,yu2024layoutdetr,hsu2025postero,tang2023layoutnuwa} focus on inferring the optimal layout of visual elements, \eg, logo, underlay, and text. However, this makes it difficult to achieve a seamless, comprehensive workflow for graphic design. To alleviate the challenge, some studies~\cite{jia2023cole,inoue2024opencole,lin2023autoposter,qu2025igd,zhang2025creatiposter,chen2025posta,kikuchi2025multimodal,li2023planning} construct an automated system to directly translate user intentions to final designs. For example, COLE~\cite{jia2023cole} introduces several task-specific models for layout planning, background\&object layers generation, and typography generation. Given the user prompt, the recent work IGD~\cite{qu2025igd} uses a unified model to generate multimodal assets and the corresponding attributes. Despite these advances, existing design systems are confined to simple design scenarios, constrained by shallow layer hierarchies and limited layer\&attribute types. Moreover, the design processes of these methods lack intuitiveness, making it difficult to simulate the creative workflow of human designers.

\myparagraph{Reinforcement Learning from Human Feedback.}
Reinforcement Learning from Human Feedback (RLHF) has proven effective in aligning model outputs with human preferences, substantially improving downstream task performance. DPO-like methods~\cite{rafailov2023direct,wallace2024diffusion} optimize the model from constructed preference pairs, aligning the model outputs with winning samples, while distancing them from losing ones. Although these methods provide a stable training process, they sometimes exhibit poor generalization ability. In contrast, PPO-like approaches~\cite{schulman2017proximal,liu2025flow} achieve better performance by optimizing the model using the policy gradient derived from estimated values. Recently, GRPO~\cite{guo2025deepseek} eliminates the dependence on an explicit value network required by traditional PPO algorithms, achieving a good balance between computation and performance.

%% file: sec/3_dataset.tex
\input{tab/dataset_compare.tex}

\section{\psddataset Dataset}
\label{sec:dataset}
\subsection{Data Construction Pipeline}
We first present \psddataset, a large-scale collection of PSD-format design files with annotated operation traces, enabling our ~\plannername to learn professional design processes from human designers.
By training on this dataset, \plannername performs designs with the following modes:
\textbf{1) } In \modeonename mode, the model harmoniously integrates the new asset into the current design. \textbf{2).} In \modetwoname mode, it refines inferior layers within a group to enhance visual quality.
To achieve this, we develop a three-stage pipeline to construct \psddataset, as illustrated in \cref{fig:dataset_pipeline}. More details are in the supplementary.

\myparagraph{Stage I: Collection of PSD Files.}
%*
As shown on the left of \cref{fig:dataset_pipeline}, we first collect a large-scale corpus of professionally designed PSD files from both the internet and paid data, encompassing diverse design scenarios and artistic styles.
%*
Next, professional annotators are employed to group related layers based on their underlying visual concepts, following design principles used by expert designers.

\myparagraph{Stage II: Parsing of PSD Files.} In this step, we parse PSD files to obtain essential information required in the subsequent stage, as illustrated in the middle of \cref{fig:dataset_pipeline}. Specifically, we first acquire the raw assets (\eg, images, texts) associated with each layer. We then extract the metadata to represent the layer hierarchy of the PSD file, where each node corresponds to either an individual layer or a layer group that encapsulates a coherent visual concept. Each node contains various attributes, such as layer type, position, opacity, blending mode, effects (\eg, inner glow, drop shadow), clipping masks, and so on. In addition, we also record the intermediate rendering results by overlaying the layers step by step.

\myparagraph{Stage III: Construction of Tool-Use Training Data.} Based on the extracted information from Stage II, we construct supervised training data for two design modes of \plannername, as shown on the right of \cref{fig:dataset_pipeline}.

\noindent\textbf{(1) Asset Integration \modeonename.}
In this mode, \plannername aims to harmoniously integrate the new asset by considering previously inserted layers within the same group, which feature high visual relevance. We randomly select layers from PSD files, and the training tuple of each layer can be constructed as:
\begin{equation}
    % (\mathcal{C}_i, a_i, x_{\text{gen},i}) = \left((M_i, R_i), a_i, x_{\text{gen},i}\right),
     ( a_\text{gen}, \mathcal{C}_\text{gen},x_{\text{gen}}) = \left(a, (M,R), x_{\text{gen}}\right),
\end{equation}
where $a_\text{gen}=a$ is the asset associated with the selected layer.
The observation $\mathcal{C}_\text{gen}$ encapsulates the layer information $M$ and current render $R$. Specifically, $M$ encompasses the necessary attributes of all preceding layers within the current group (\eg, layer type, position), which can be obtained from the extracted metadata.
$R$ can be retrieved from the recorded intermediate renders.
By integrating $a$, the constructed tool calls $x_\text{gen}$ can faithfully convert the current render $R$ to the recorded subsequent one, as indicated in the right of \cref{fig:dataset_pipeline}. In particular, we use predefined rules to convert the metadata of the selected layer into $x_\text{gen}$, as demonstrated in the supplementary.
Through learning to predict $x_\text{gen}$ based on $(a,(M,R))$, \plannername learns to integrate the new asset harmoniously.

\noindent\textbf{(2) Layer Refinement \modetwoname.}
For this mode, \plannername aims to refine inferior layers within the current group, thereby composing an appealing visual concept. The training tuple for each randomly selected layer is constructed as:
\begin{equation}
    (a_\text{edt},\mathcal{C}_\text{edt}, x_{\text{edt}}) = \big(\emptyset,(M,R, G), x_{\text{edt}}\big),
\end{equation}
where $a_\text{edt}=\emptyset$ indicates there is no asset for this mode. For $\mathcal{C}_\text{edt}$, $M$ is constructed upon distorted metadata, which modifies the attributes of the selected layer or its preceding layers within the same group (such as position, opacity, \etc). Then, the current render $R$ is obtained by modifying the PSD file based on $M$, followed by a re-rendering process, as indicated on the right of \cref{fig:dataset_pipeline}. Furthermore, we also incorporate $G$ into the current observation, which is the rendered image before the group is applied. The visual differences between $R$ and $G$ provide crucial cues for identifying inferior layers. We can easily retrieve $G$ from the recorded renders in Stage II since the layers that precede the group remain unchanged.
The refinement tool calls $x_{\text{edt}}$ can also be derived by the predefined rules introduced in the supplementary, which recover the original layer configuration, as shown on the right of \cref{fig:dataset_pipeline}.
By learning to predict $x_\text{edt}$ based on $(M,R, G)$, \plannername learns to detect and refine the inferior layers.

In addition, we also need to construct positive training samples to prevent the \plannername from inferring unnecessary refinement tool calls under optimal configuration. To this end, we construct $\mathcal{C}_\text{edt}$ from the original metadata and recorded renders, while $x_{\text{edt}}$ is set to empty.

\subsection{Comparison with Existing Datasets}
To highlight the strengths of \psddataset, we compare it with existing graphic design datasets in \cref{tab:datasets}.
Benefiting from the professional structure of the PSD format, our dataset offers several notable advantages:
\textbf{1) Complex layer hierarchies.} \psddataset, with an average of approximately 48.35 layers per sample, provides far more complex layer organizations than previous datasets, allowing models to learn from challenging and realistic design compositions.
\textbf{2) Diverse layer and attribute types.} Prior datasets are limited by their file formats to mainly simple layer types (such as images and text), and a relatively small set of attributes (like position and font). In contrast, \psddataset covers a substantially wider range of layer types, like adjustment layers, and a richer set of attributes, such as blending modes, layer effects, and clipping masks. This diversity enables more expressive modeling of design operations and better represents real-world design workflows.
\textbf{3) Intuitive grouping strategy.} Layers in our dataset are grouped by visual concept, forming organized hierarchies that align with the design principle of experts.
These features make \psddataset a more effective resource than previous datasets for training models to learn human-like creative design workflows, perform a wide variety of operations, and tackle complex compositional tasks.

% TODO 46.6

%% file: tab/dataset_compare.tex
\begin{table*}[t]
\centering
\small
\definecolor{mygray}{gray}{0.99}
\caption{{Representative graphic design datasets. * denotes inaccessible fields.}}
\vspace{-3mm}
\begin{threeparttable}
\scriptsize
\setlength{\tabcolsep}{7.6pt}
\renewcommand{\arraystretch}{1.07}
\resizebox{1\textwidth}{!}{%
\begin{tabular}{l|r|c|c|l}
\toprule
\textbf{Dataset} & \textbf{\#Samples} & \textbf{\#Layers} & \textbf{Layer Types} & \ \ \ \ \ \ \textbf{Attribute Types} \\
\hline
\rowcolor{myrowgray!50} CGL~\cite{cgl-pami} & 60,548 & \ \ {4.80} & image, text (2) & bbox, category...(7) \\
Crello~\cite{yamaguchi2021canvasvae} & 23,182 & \ \ {4.29} & image, text (2) & type, opacity...(33)  \\
\rowcolor{myrowgray!50} Design39K~\cite{design39k} & 39,725 & \ \ 5.13 & image, text (2) & type, font...(*) \\
% IGD~\cite{qu2025igd} & 92,000 & \ \ 1.00 & 2 & position, color...(*) \\
AutoPoster~\cite{lin2023autoposter} & \textbf{76,960} & \ \ {5.43} & image, text (2) & position, size...(*) \\
% \rowcolor{myrowgray!50}PPG30k~\cite{ppg30k} & 34,150 & \ \ 1.00 & image, text (2) & category, width...(5)  & No\\
% \rowcolor{myrowgray!50} PosterSum~\cite{paper2poster} & 16,305 & \ \ 1.00 & 2 & topics...(7) \\
\rowcolor{myrowgray!50} PKU PosterLayout~\cite{hsu2023posterlayout} & 9,974 & \ \ {4.53} & image, text (2) & bbox, type (2)\\
% \rowcolor{myrowgray!50} PosterArt~\cite{chen2025posta} & $>$2,000 & \ \ \textcolor{red}{3.82} & image, text (2) & bbox, color...(6) & No\\
% PosterCraft~\cite{chen2025postercraft} & \textbf{$>$2,000,000} & \ \ 1.00 & image, text (2) & position, orientation...(*) & No\\
\rowcolor{cyan!10}
\textbf{\psddataset~(Ours)} & 10,454 & \textbf{48.35} & \textbf{smart object, adjustment... (5)}~ & \textbf{effects, opacity... ($>$60)} \\
\bottomrule
\end{tabular}}
\end{threeparttable}
\label{tab:datasets}
% \vspace{-3mm}
\end{table*}

%% file: sec/3_method.tex
\section{Method: \systemname}
\label{sec:method}
{To emulate the creative and innovative workflow of human designers, we construct \systemname, an automated graphic design system that translates user intentions into PSD-format design files through multiple coordinated components.
The overall design workflow is illustrated in \cref{sec:workflow}.
In particular, \collectorname (\cref{sec:collector}) first collects theme-related assets based on user instructions.
Then, \plannername (\cref{sec:planner}), trained on our curated design dataset \psddataset, predicts tool calls based on the current design.
Finally, \executorname (\cref{sec:executor}) performs these tool calls to manipulate the PSD file.
}

\subsection{Design Workflow of \systemname}\label{sec:workflow}
The bottom of \cref{fig:teaser} shows a graphic design workflow of \systemname. First, \collectorname identifies the visual concepts from user instructions and collects their related assets. 
Then, the system is ready to iteratively integrate these assets into the design, where a bottom-up traversal is performed on the nested hierarchy, first at the group level and then at the asset
level. We now introduce the procedure of each iteration.

For the sake of clarity, we first denote $i$ as the current index,  which is incremented by one following each manipulation of the design file. \systemname performs the following step for each iteration: \textbf{1).} Based on the current asset $a_i$ and the observation $\mathcal{C}_{\text{gen},i}=(M_i,R_i)$, \plannername first works on \modeonename mode to predict the tool calls $x_{\text{gen},i}$ for asset integration. \textbf{2).} Then, \executorname performs $x_{\text{gen},i}$ to harmoniously incorporate the asset into the design file, and the index becomes ${i+1}$. \textbf{3).} Next, \plannername that works on \modetwoname mode infers the tool calls $x_{\text{edt},i+1}$ to refine the inferior layers based on the observation $\mathcal{C}_{\text{edt},i+1}=(M_{i+1},R_{i+1},G_{i+1})$. \textbf{4).} \executorname performs $x_{\text{edt},i+1}$ to retouch the design, and the index becomes ${i+2}$.
The above steps are repeated until all assets are integrated into the design file.

\subsection{\collectorname}\label{sec:collector}
As indicated on the lower left of \cref{fig:teaser}, given a user instruction, \collectorname first identifies the possible visual concepts. Specifically, it leverages a pretrained Large Language Model (LLM) to achieve the goal through a well-designed query prompt. Then, it collects related assets for each concept. In particular, the image assets are sourced from the internet, databases, or image generation models, while the textual assets can be directly derived from the used LLM. For some artistic fonts, \collectorname also leverages image generation models to generate stylized text images for conveying textual information.
More details are provided in the supplementary.

\begin{figure*}[t]
    \centering
    \includegraphics[width=1\linewidth]{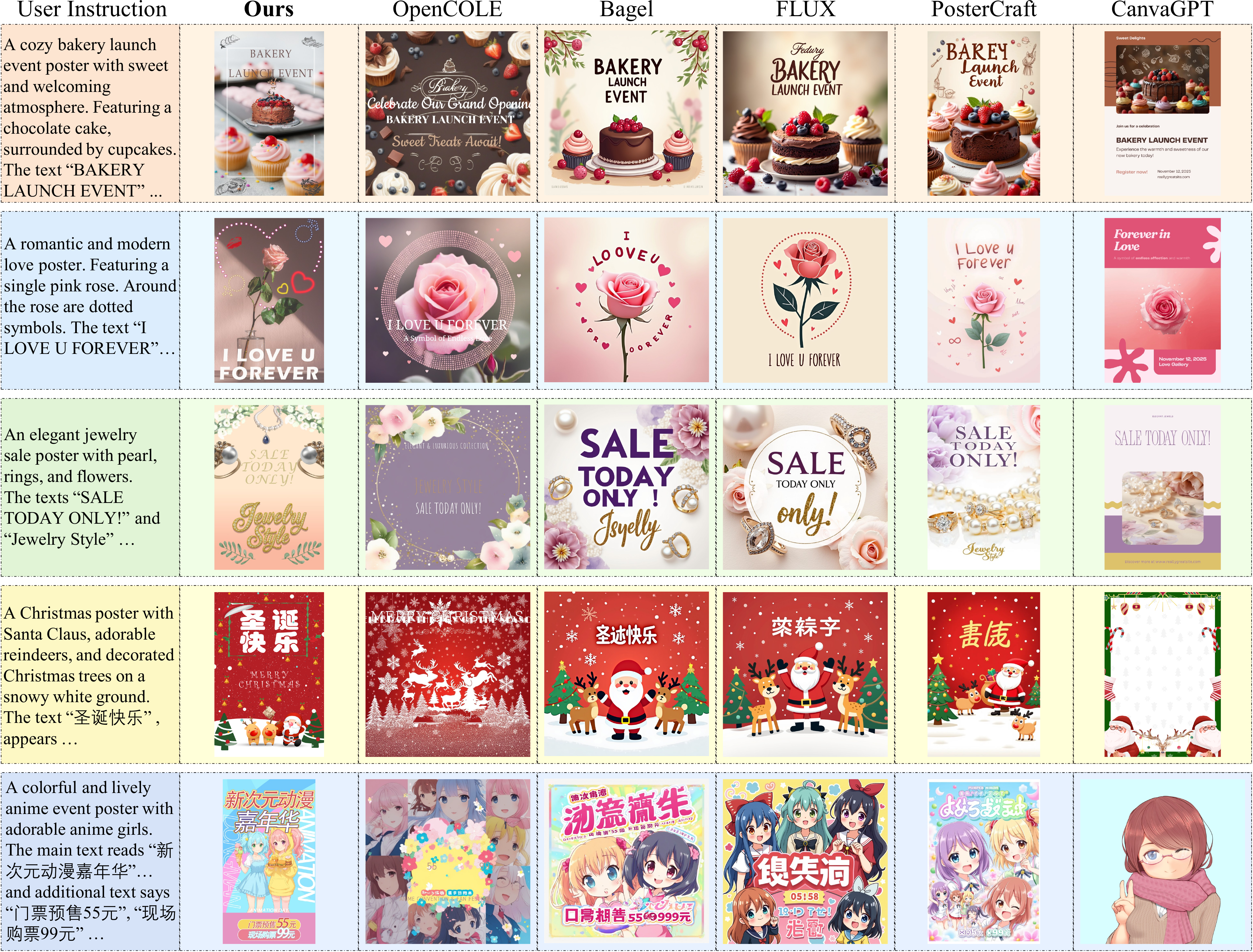}
    \vspace{-6.6mm}
    \caption{Evaluation of the model performance on translating user intentions to the final designs. Most of the compared methods can only generate non-editable raster images or output a few layers with simple attributes, hindering their professionalism and flexibility. Furthermore, many methods can not generate accurate texts, especially for complex characters, \eg, Chinese.}
   \label{fig:exp1}
   \vspace{-3mm}
\end{figure*}

\subsection{\plannername}\label{sec:planner}
As mentioned in \cref{sec:dataset}, \plannername predicts the tool calls and performs in two design modes: \modeonename for incorporating new assets and \modetwoname for refining the inferior layers. By using the curated design dataset \psddataset, we train \plannername in two stages to equip it with strong tool-use capabilities.

\myparagraph{Supervised Fine-Tuning Stage.} To process the multimodal inputs, we build the \plannername upon the pre-trained VLM~\cite{wang2024qwen2}. In training, we inject mode-specific LoRA modules and learn \plannername with the following objective:
\begin{equation}\label{eq:sft}
\mathcal{L} = - \mathbb{E}_{(a,\mathcal{C},x) \sim \mathcal{D_\text{gen/edt}}} \sum_{t=1} \log p_{\theta_\text{gen/edt}}(x_t\mid x_{<t},a,\mathcal{C}),
\end{equation}
where the asset $a$, observation $\mathcal{C}$, and the tool calls $x$ are sampled from mode-specific training dataset $D_\text{gen/edt}$ mentioned in \cref{sec:dataset}. Specified by the LoRA weights $\theta_\text{gen}$ and $\theta_\text{edt}$, $p_{\theta_\text{gen/edt}}$ indicates the \plannername under \modeonename and \modetwoname modes, respectively.

\myparagraph{Reinforcement Learning Stage.} We apply GRPO~\cite{guo2025deepseek} to enhance the model’s tool-use proficiency. Given the condition $(a,\mathcal{C})$, the designed reward function $r$ compares the generated tool calls with ground truth, while rewarding the samples if the tool names and the corresponding parameter names\&values are correctly predicted.  We provide the formulation of $r$ in the supplementary. Then, we optimize \plannername to maximize the GRPO objective:
{\small
\begin{align}
&\mathbb{E}_{(a,\mathcal{C}) \sim D,\, \{x_i\}_{i=1}^{N_G} \sim p_{{\text{ref}}}(x|a,\mathcal{C})}
\Bigg[\frac{1}{N_G} \sum_{i=1}^{N_G} \Big(\min \Big(\frac{p_\theta(x_i|a,\mathcal{C})}{p_{\text{ref}}(x_i|a,\mathcal{C})}A_i,\\
&\nonumber\text{clip}\Big(\frac{p_\theta(x_i|a,\mathcal{C})}{p_{{\text{ref}}}(x_i|a,\mathcal{C})},1-\epsilon,\,1+\epsilon\Big)A_i\Big)- \beta\, \mathbb{D}_{\text{KL}}\!\left(p_\theta \,\|\, p_{\text{ref}}\right)\Big)\Bigg],
\end{align}
}where $p_\text{ref}$ indicates the model from the supervised fine-tuning stage, and the mode subscripts $\text{gen}/\text{edt}$ are omitted for simplicity. $N_G$ is the group size and $\epsilon$ is the clipping hyperparameter. $A_i$ is calculated as ${(r_i - \mathrm{mean}(\{r_1, r_2, \cdots, r_{N_G}\}))}/{\mathrm{std}(\{r_1, r_2, \cdots, r_{N_G}\})}$. $\beta$ is the weight of KL-divergence $\mathbb{D}_{\text{KL}}$ for regularization.
% }

\subsection{\executorname}\label{sec:executor}
We introduce \executorname to enable seamless collaboration between \systemname and Adobe Photoshop, performing the predicted tool calls to manipulate the PSD file. In particular, we implement \executorname based on Unified Extensibility Platform (UXP), which is Adobe’s modern extension framework that empowers building and running JavaScript APIs inside Photoshop. 
As a result, our implementation includes over 70 tools, such as operations for inserting the image, text, or adjustment layers, as well as applying effects (such as inner glow or drop shadow).
More details can be found in the supplementary.

% TODO 70, 

%% file: sec/4_experiments.tex
\begin{figure*}[!t]
    \centering
    \includegraphics[width=1\linewidth]{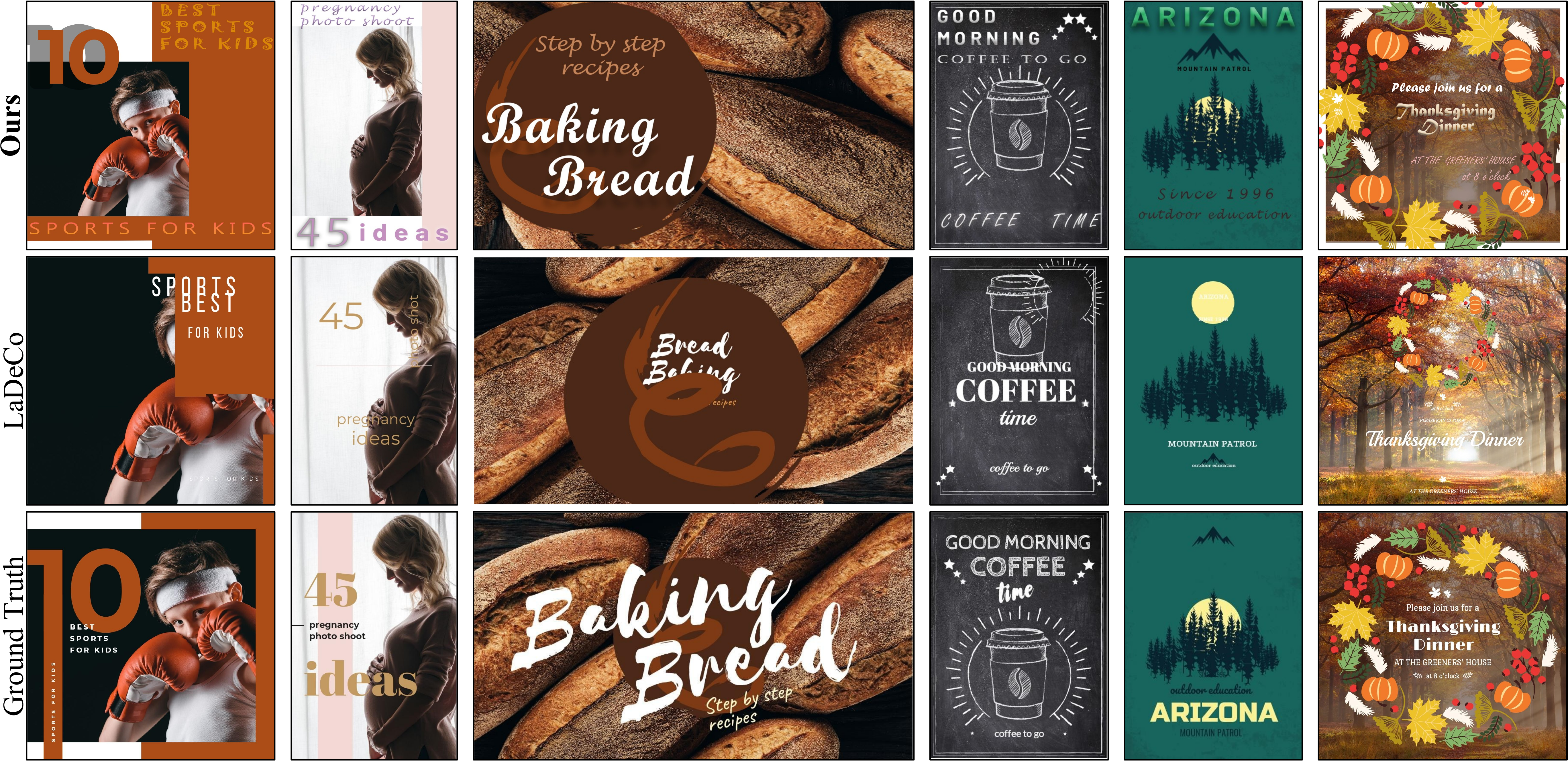}
    \vspace{-6mm}
    \caption{Evaluation of model performance on the graphic design composition task, using Crello-v5 dataset~\cite{yamaguchi2021canvasvae}. Our method achieves coherent arrangements, achieving visually appealing outcomes.}
   \label{fig:exp2}
\end{figure*}

\section{Experiments}
\label{sec:experiments}
\subsection{Experimental Setup}
\myparagraph{Implementation Details.}
Our \plannername is implemented based on Qwen2.5-VL-7B~\cite{wang2024qwen2}. All experiments are conducted on 4 NVIDIA A800 80G GPUs.
During the supervised fine-tuning stage of \plannername, we introduce distinct LoRA modules for \modeonename and \modetwoname modes, which are learned on the respective training dataset with a batch size of 64 and a learning rate of 2e-4, for 15,000 and 12,000 steps, respectively. The rank numbers of introduced LoRAs are all set to 32. In the reinforcement learning stage, we optimize \plannername with the GRPO objective for 6,000 steps, and the group size is set to 8. The dataset of this stage is derived from 4,000 PSD files of \psddataset.

\myparagraph{Evaluation Details.} 
To demonstrate the effectiveness of our method, we conduct the following experiments. \textbf{1).} We first evaluate the model’s ability to directly translate user intentions into final designs. Similar to previous works~\cite{qu2025igd}, we collect 250 user instructions for both English and Chinese scenarios. The compared methods include open-source models like OpenCOLE~\cite{inoue2024opencole}, Bagel~\cite{deng2025emerging}, FLUX~\cite{black2024flux}, PosterCraft~\cite{chen2025postercraft}, and commercial model CanvaGPT.
\textbf{2).} We further assess the model’s capability to perform graphic design composition based on the given assets. Specifically, we use the test data from Crello-v5~\cite{yamaguchi2021canvasvae} to evaluate the model performance in simple design scenarios. We compare our method with the advanced model LaDeCo~\cite{lin2025elements}, which can process multimodal inputs and predict the layer attributes. We further evaluate our method on 200 copyright-free PSD files as a complement, featuring complex layer hierarchies.

Following previous works~\cite{jia2023cole,lin2025elements,qu2025igd,inoue2024opencole}, we employ VLM~\cite{wang2024qwen2} to evaluate designs across the following aspects: aesthetic quality (\textbf{Qua.}), design layout (\textbf{Lay.}), content relevance (\textbf{Rel.}), color harmony (\textbf{Har.}), and innovation(\textbf{Inn.}). Moreover, we also conduct a user study to evaluate these dimensions. All scores are within the range of 1 to 10. Please kindly refer to the supplementary for more details.

\input{tab/quanti_compare}

\subsection{Comparison on Graphic Design}
We first evaluate the model’s performance to directly translate user intentions into final designs. It is worth noting that our approach presents a more challenging design process compared to other methods, as it requires harmoniously arranging multiple multimodal assets and inferring complex layer attributes. As shown in \cref{fig:exp1}, most of the methods can generate high-quality visual outcomes. However, except for our method and OpenCOLE~\cite{inoue2024opencole}, other models produce non-editable raster images, hindering the flexibility in adding customized assets or refining the content. Furthermore, the outputs from OpenCOLE only contain a single image layer and several text layers with simple attributes, constraining its editability. In contrast, our \systemname produces PSD-format design files with complex layer hierarchies, achieving higher professionalism and flexibility. Moreover, the performance of the compared methods is constrained by their text rendering abilities, resulting in distorted/extraneous/missing characters. This is pronounced with complex characters (\eg, Chinese), as shown in the last two rows of \cref{fig:exp1}, hindering them from creating accurate texts. Our method and OpenCOLE address this issue by harmoniously overlaying the text layers. As shown in \cref{tab:exp1}, our \systemname achieves competitive quality with other advanced methods across most evaluation dimensions.

Next, we assess the model’s ability to perform graphic design composition based on the given assets. As shown in \cref{fig:exp2}, LaDeCo~\cite{lin2025elements} is prone to generating inaccurate layouts, resulting in occlusion of key subjects (1st and 3rd columns of \cref{fig:exp2}) or suboptimal placement of elements (4th and 5th columns). In contrast, our \plannername produces coherent arrangements, resulting in visually appealing outcomes. Moreover, \plannername also adds effects (\eg, shadows) to enhance the harmony of the design. \cref{tab:exp2} further demonstrates that our method outperforms in most of the evaluation dimensions.

\begin{figure}[t]
    \centering
    \includegraphics[width=1\linewidth]{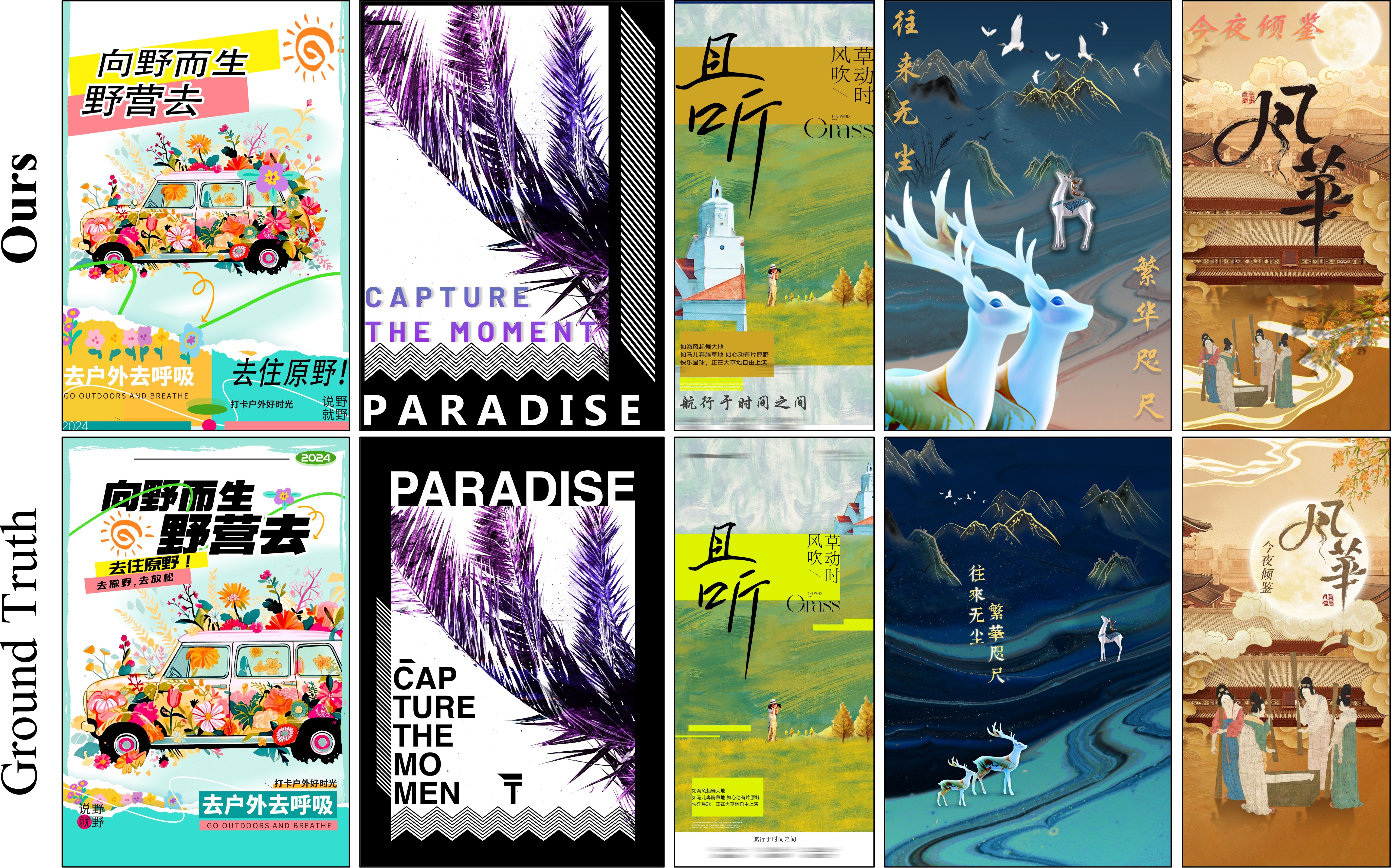}
    \vspace{-6mm}
    \caption{The results of our method in graphic design composition tasks with complex layer hierarchies.}
   \label{fig:exp3}
\end{figure}

Constrained by the training dataset~\cite{yamaguchi2021canvasvae}, LaDeCo struggles with handling complex layer hierarchies. Consequently, \cref{fig:exp3} and \cref{tab:ablation} exclusively demonstrate our method's performance in composing the assets from collected PSD files, highlighting its superior capability in handling challenging design scenarios.

\input{tab/quanti_compare2}

\input{tab/quanti_ablation}
\subsection{Ablation Study}

\cref{tab:ablation} shows the results of ablation studies. We compare our \plannername with the following settings: without \modetwoname mode (w/o \modetwoname), without layer information $M$ (w/o $M$), and without reinforcement learning (w/o RL). As shown in \cref{tab:ablation}, models under these settings exhibit inferior performance across all evaluation dimensions. For the previous two settings, \plannername lacks sufficient awareness of other elements within the current group, leading to inaccurate prediction of the corresponding layer attributes. In addition, without reinforcement learning, the method fails to predict precise parameter values within the tool calls, ultimately leading to suboptimal visual outcomes. More details are provided in the supplementary.

%% file: tab/quanti_compare.tex
\begin{table*}[t]
\centering
\footnotesize
% \caption{The quantitative results evaluated by the VLM~\cite{bai2025qwen2}, are presented using two metrics: VLM (1 - Normalized Edit Distance) and Sentence Accuracy (Human), which is 1 if the entire sentence is correct and 0 otherwise. The 'English-Thai' and 'Complex' categories show the performance on the \multilingualbenchmarkname and \complexbenchmarkname benchmarks, respectively.}
\caption{VLM and human evaluation scores for graphic design. All scores are within the range of 1 to 10.}
\vspace{-2mm}
\label{tab:exp1}
% \scriptsize
\setlength{\tabcolsep}{6pt}
\renewcommand{\arraystretch}{1.06}
\resizebox{\textwidth}{!}{ % Resizes the table to fit the text width
\begin{tabular}{c|cc|cc|cc|cc|cc|cc}
\toprule
\textbf{Metric} & \multicolumn{2}{c|}{\textbf{Ours}} & \multicolumn{2}{c|}{OpenCOLE~\cite{inoue2024opencole}} & \multicolumn{2}{c|}{Bagel~\cite{deng2025emerging}} & \multicolumn{2}{c|}{FLUX~\cite{black2024flux}} & \multicolumn{2}{c|}{PosterCraft~\cite{chen2025postercraft}} & \multicolumn{2}{c}{CanvaGPT} \\
\cline{2-13}
 & \textbf{VLM} & \textbf{Human} & \textbf{VLM} & \textbf{Human} & \textbf{VLM} & \textbf{Human} & \textbf{VLM} & \textbf{Human} & \textbf{VLM} & \textbf{Human} & \textbf{VLM} & \textbf{Human} \\
\hline
\textbf{Qua.} & 7.62 & 8.09 & 5.12 & 3.95 & 5.85 & 6.12 & \underline{8.18} & 7.68 & 7.95 & \textbf{8.35} & \textbf{8.52} & \underline{8.28} \\
\textbf{Lay.} & \textbf{8.68} & \textbf{9.12} & 6.85 & 4.58 & 4.47 & 6.88 & 3.66 & 5.41 & 6.39 & \underline{8.15} & \underline{8.35} & 7.08 \\
\textbf{Rel.} & \underline{7.78} & \underline{7.17} & 5.25 & 3.48 & 6.58 & 5.24 & 6.92 &\textbf{8.15} & \textbf{8.42} & 7.10 & 4.72 & 6.88  \\
\textbf{Har.} & \underline{8.02} & \underline{7.72} & 6.68 & 5.92 & 7.25 & 6.08 & 6.82 & 5.02 & \textbf{8.05} & 7.65 & 7.21 & \textbf{7.88} \\
\textbf{Inn.} & \textbf{8.45} & {6.98} & 6.08 & \underline{7.32} & 5.68 & 6.45 & 6.95 & {7.18} & 5.87 & \textbf{7.58} & \underline{7.52} & 6.33 \\
\bottomrule
\end{tabular}
}
% \vspace{-3mm}
\end{table*}

% \begin{table}[h]
% \centering
% \begin{tabular}{l|cc|cc|cc|cc|cc|cc}
% \textbf{Metric} & \multicolumn{2}{c|}{\textbf{PSDesigner (Ours)}} & \multicolumn{2}{c|}{\textbf{OpenCOLE}} & \multicolumn{2}{c|}{\textbf{Bagel}} & \multicolumn{2}{c|}{\textbf{FLUX}} & \multicolumn{2}{c|}{\textbf{PosterCraft}} & \multicolumn{2}{c}{\textbf{CanvaGPT}} \\
% & \textbf{VLM} & \textbf{Human} & \textbf{VLM} & \textbf{Human} & \textbf{VLM} & \textbf{Human} & \textbf{VLM} & \textbf{Human} & \textbf{VLM} & \textbf{Human} & \textbf{VLM} & \textbf{Human} \\
% \hline

% \end{tabular}
% \caption{}
% \end{table}

%% file: tab/quanti_compare2.tex
\begin{table}[t!]
\centering
\small
\setlength{\tabcolsep}{11pt}
\caption{VLM evaluation scores for graphic design composition, using Crello-v5~\cite{yamaguchi2021canvasvae} dataset.}
\vspace{-3mm}
\renewcommand{\arraystretch}{1.03}
% \resizebox{0.37\textwidth}{!}{
\begin{tabular}{r|cccc}
\toprule
{Method} & \textbf{Qua.} & \textbf{Lay.} & \textbf{Har.} & \textbf{Inn.} \\
\hline
\textbf{Ours} & \textbf{7.85} & \textbf{7.43} & {6.77}  & \textbf{6.94} \\
LaDeCo~\cite{lin2025elements} & {5.95} & {6.03} & \textbf{7.22}& {5.75} \\
\hline
Ground Truth & 8.13 & 9.18 & 8.90 & 7.12 \\
\bottomrule
\end{tabular}
% }
\label{tab:exp2}
\end{table}

%% file: tab/quanti_ablation.tex
\begin{table}[t!]
\centering
\small
\setlength{\tabcolsep}{8.5pt}
\renewcommand{\arraystretch}{1.03}
\caption{Ablation studies.}
\vspace{-3mm}
% \renewcommand{\arraystretch}{1.1}
% \resizebox{0.40\textwidth}{!}{ 
\begin{tabular}{c|l|cccc}
\toprule
{} &{Method} & \textbf{Qua.} & \textbf{Lay.} & \textbf{Har.} & \textbf{Inn.} \\
\hline
\multirow{4}{*}{\rotatebox[origin=c]{90}{Crello-v5}}
    & \textbf{Ours} & \textbf{7.85} & \textbf{7.43} & \textbf{6.77}  & \textbf{6.94} \\
    & {Ours w/o \modetwoname} & 6.05 & 5.88 & 5.90 & \underline{6.75} \\
    & {Ours w/o $M$} & 6.25 & \underline{6.10} & 6.18 & 6.02 \\
    & {Ours w/o RL} & \underline{6.38} & 6.00 & \underline{6.35} & {6.20} \\
\hline
\multirow{4}{*}{\rotatebox[origin=c]{90}{PSD}}
    & \textbf{Ours} & \textbf{6.28} & \textbf{6.15} & \textbf{7.02} & \textbf{6.88} \\
    & {Ours w/o \modetwoname} & 5.32 & 5.15 & 6.22 & 6.05 \\
    & {Ours w/o $M$} & 5.60 & 5.38 & 6.52 & 6.35 \\
    & {Ours w/o RL} & \underline{5.72} & \underline{5.48} & \underline{6.68} & \underline{6.57} \\
\bottomrule
\end{tabular}
% }
\label{tab:ablation}
\end{table}

%% file: sec/5_conclusion.tex
\section{Conclusion}
\label{sec:conclusion}
Graphic design is a creative yet expertise-intensive process that often requires substantial professional skills and manual effort, making it inaccessible to non-specialists. To lower this barrier, we propose \systemname, an automated design system that translates user intentions into editable PSD-format files. Given a user instruction, \collectorname gathers relevant assets for each identified visual concept. Then, \plannername and \executorname collaboratively infer and execute tool calls to integrate assets or refine suboptimal layers. To endow \plannername with strong tool-use capabilities, we construct a large-scale dataset, \psddataset, derived from PSD files annotated with detailed operation traces. By training on \psddataset, the model learns expert-level design workflows and supports a wide range of visual editing tasks. Extensive experiments validate the effectiveness of our system in enabling non-specialists to generate production-quality graphic designs.